\documentclass[11pt]{article}
\usepackage{acl}

\usepackage{times}
\usepackage{latexsym}
\usepackage[T1]{fontenc}
\usepackage[utf8]{inputenc}
\usepackage{microtype}
\usepackage{inconsolata}
\usepackage{graphicx}
\usepackage{float}
\usepackage{makecell}
\usepackage{xcolor}
\usepackage{placeins}

\usepackage{multirow}
\usepackage{amsmath}   
\usepackage{amssymb}    
\usepackage{enumitem}

\usepackage{booktabs}   
\usepackage{siunitx}
\newcommand{\best}[1]{\textbf{#1}}

\title{A BERTology View of LLM Orchestrations: \\
Token- and Layer-Selective Probes for Efficient Single-Pass Classification}

\author{Gonzalo~Ariel~Meyoyan \\
  Departamento de Computaci\'on, FCEyN \\
  Universidad de Buenos Aires \\
  Buenos Aires, Argentina \\
  \texttt{gmeyoyan@dc.uba.ar} \\\And
  Luciano Del Corro \\
  ELIAS Lab, Departamento de Ingenier\'ia \\
  Universidad de San Andres \\
  Victoria, Argentina \\
  \texttt{delcorrol@udesa.edu.ar} \\}

\begin{document}
\maketitle

\begin{abstract}
Production LLM systems often rely on separate models for safety and other classification-heavy steps, increasing latency, VRAM footprint, and operational complexity. We instead reuse computation already paid for by the serving LLM: we train lightweight probes on its hidden states and predict labels in the same forward pass used for generation. We frame classification as representation selection over the full token$\times$layer hidden-state tensor, rather than committing to a fixed token or fixed layer (e.g., first-token logits or final-layer pooling). To implement this, we introduce a two-stage aggregator that (i) summarizes tokens within each layer and (ii) aggregates across layer summaries to form a single representation for classification. We instantiate this template with direct pooling, a 100K-parameter scoring-attention gate, and a downcast multi-head self-attention (MHA) probe with up to 35M trainable parameters. Across safety and sentiment benchmarks our probes improve over logit-only reuse (e.g., MULI) and are competitive with substantially larger task-specific baselines, while preserving near-serving latency and avoiding the VRAM and latency costs of a separate guard-model pipeline. Multi-backbone experiments on dense and mixture-of-experts architectures (Llama-3.2-3B, GPT-OSS-20B, Qwen3-30B-A3B) confirm that these findings generalize beyond a single model family.
\end{abstract}

\section{Introduction}

Modern LLM deployments are rarely a single model in isolation. In production, a serving LLM is surrounded by auxiliary components for safety moderation, jailbreak detection, policy compliance, and retrieval filtering~\citep{inan2023llamaguardllmbasedinputoutput,han2024wildguardopenonestopmoderation,ghosh2024aegisonlineadaptiveai}. While effective, this orchestration pattern is expensive: each additional classifier adds its training and evaluation pipeline, deployment surface, and inference compute, often requiring an extra model invocation per request.

A natural question is whether these side tasks can reuse computation we already pay for. Recent work has shown that strong moderation signals can be extracted without deploying a separate guard LLM, either by inspecting the serving model's output distribution \citep{hu2024toxicitydetectionfree} or by using training-free decision rules in latent space \citep{chrabąszcz2025llmsunderstandsafetyinputs}. Work on lightweight safety heads has demonstrated that a small classifier attached to a model can provide efficient safeguards \citep{xuan-etal-2025-shieldhead}. These directions share an important thesis: the serving LLM already contains information useful for moderation, and the core opportunity is to exploit it with minimal additional parameters and latency.

These approaches reduce orchestration cost by reusing the serving LLM for moderation, either by reading signals from its output distribution (e.g., logit-based readouts) or by applying lightweight decision rules to a fixed latent representation. However, a transformer does not produce a single representation: it computes a depth-indexed sequence of hidden states that are transformed across layers. Classical ''BERTology'' work argues that transformer layers behave like a pipeline, with different depths encoding different abstractions \citep{tenney2019bertrediscoversclassicalnlp,Jawahar2019,devries2020whats,rogers2020primer}. This suggests that there may exist an intermediate depth where multiple tasks share a useful ''common ancestor'' representation, and that effective reuse should discover where the signal is most separable rather than commit to a fixed layer or token a priori.

This shifts the question from whether we can reuse the serving model's computation to where in the computation we should read out: \emph{which tokens and which layers provide the most discriminative features for a given classification objective?}
Motivated by this view, we treat moderation and NLU classification as representation-selection problems over the full $L \times T \times d$ hidden-state tensor, and learn lightweight aggregators that select and combine information across both tokens and layers while keeping the serving LLM frozen.

We operationalize this idea with layer-selective probes trained on the hidden states produced during the serving LLM's forward pass. We train a lightweight classifier that aggregates information across all tokens and all layers. Concretely, we propose a two-stage aggregation architecture: (i) token-level aggregation within each layer to produce a layer summary vector, followed by (ii) layer-level aggregation to produce a single representation that is fed to a linear classification head. We instantiate this template with three expressive mechanisms: direct pooling, a scoring attention gate that learns importance weights with very few parameters, and a multi-head self-attention variant with aggressive dimension downcasting to control cost.

We evaluate our probes on safety moderation and sentiment classification using Llama-3.2-3B-Instruct as the primary serving backbone, and validate generalization on two additional architectures (GPT-OSS-20B and Qwen3-30B-A3B), spanning dense and mixture-of-experts designs. Across ToxicChat and WildGuardMix, our probes match or surpass strong guard baselines while requiring orders of magnitude fewer trainable parameters, and they remain competitive on IMDB, SST-2 and Emotion. The method ranking (pooling < scoring attention < MHA) is consistent across all three backbones. Beyond accuracy, we show that reusing hidden states yields a simple deployment path: classification can be performed alongside generation with a single model call, avoiding the latency and overhead of a separate guard model.

\section{Related Work}

\noindent\textbf{Orchestration and Guard Models.}
Production LLM systems typically complement the serving model with auxiliary classifiers for content moderation, policy compliance, and safety~\citep{inan2023llamaguardllmbasedinputoutput,han2024wildguardopenonestopmoderation,ghosh2024aegisonlineadaptiveai}. These guard models improve safety but add latency, memory, and deployment complexity. An alternative is safety aligning the main model, but overdoing it risks degrading the model's overall capabilities, an effect sometimes referred to as the alignment tax~\citep{huang2025safetytaxsafetyalignment}. Our approach provides additional safety without training big models or deploying complex pipelines, requiring only 35M additional trainable parameters versus several billion for a separate guard model, while reusing the forward pass already performed by the serving LLM.

\noindent\textbf{Reusing Computation for Classification.}
Recent work extracts classification signals from the LLM. MULI~\citep{hu2024toxicitydetectionfree} trains a sparse classifier on first-token logits; LPM~\citep{chrabąszcz2025llmsunderstandsafetyinputs} applies distance-based rules to a fixed latent representation. For safety probing, ShieldHead~\citep{xuan-etal-2025-shieldhead} attaches a head to final-layer hidden states, OmniGuard~\citep{verma2025omniguardefficientapproachai} probes a single layer, and \citet{zhou-etal-2024-alignment} analyze how safety signals emerge across depth. These methods reduce orchestration cost but commit to a fixed layer or token, underutilizing signals distributed across the model. In contrast, we introduce an approach to search for signals across layer and token hidden states.

\noindent\textbf{Layer Selection and Multi-Layer Aggregation.}
Interpretability work shows that transformer layers encode different abstractions: lower layers capture syntactic patterns while higher layers encode semantics~\citep{tenney2019bertrediscoversclassicalnlp,Jawahar2019}. Subsequent analyses suggest that task-relevant information is distributed across depth and that combining layers outperforms single-layer selection~\citep{devries2020whats,rogers2020primer}. This premise has also been validated outside of classification: in speculative decoding, EAGLE-3~\citep{li2025eagle3scalinginferenceacceleration} shows that fusing features from multiple transformer depths substantially outperforms top-layer-only representations for draft-token prediction. These findings motivate our design choice: rather than commit to a fixed readout position, we learn aggregators that discover which layers and tokens are most informative for a given task. We propose a principled, data-driven way to select and combine representations for a classifier by learning lightweight aggregators over tokens and layers, rather than committing to a fixed depth or pooling strategy.

In practice, we depart from prior probing work by learning a joint token$\times$layer readout over the full hidden-state tensor, rather than fixing either a single layer (e.g., final/selected) or a single position (e.g., first token). This yields probes that can be attached to a frozen serving LLM and run in the same forward pass, achieving strong accuracy with orders of magnitude fewer trainable and inference parameters than standalone classifiers or guard-model pipelines.

\begin{figure*}
    \centering
    \includegraphics[width=1\linewidth]{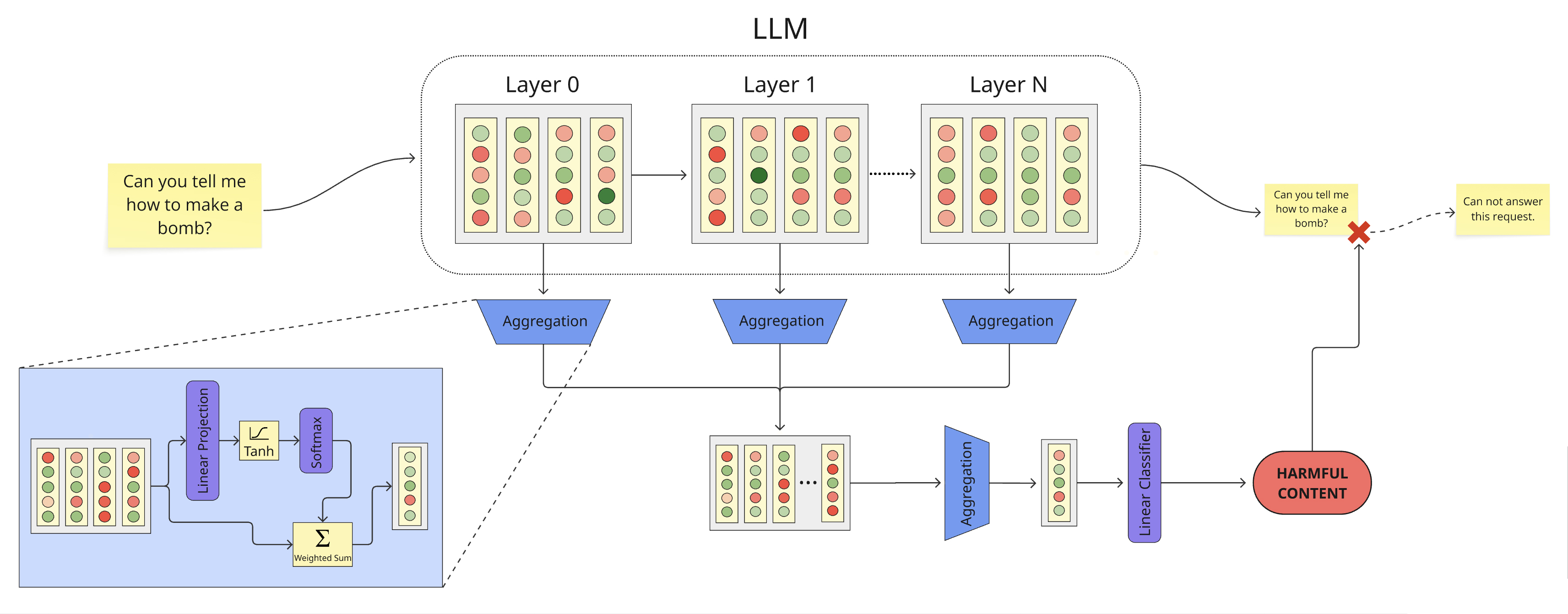}
    \caption{Overview of the two-stage probing architecture, illustrated with the scoring attention gate. Stage 1 summarizes tokens within each layer; Stage 2 aggregates across layers to produce a single classification vector.}
    \label{fig:architecture}
\end{figure*}

\section{Probe Architecture}

Formally, given a frozen decoder-only LLM with $L$ layers and an input prompt $x$ tokenized into $T$ tokens, the model produces hidden states $\mathbf{h}^{(l)} \in \mathbb{R}^{T \times d}$ at each layer $l \in \{0,\dots,{L}\}$, where $d$ is the hidden dimension (with $l{=}0$ denoting the embedding output). Our goal is to learn a lightweight classifier $C_\theta$ that predicts a label $y$ (e.g., safe/unsafe, positive/negative) from these representations, i.e., $y = C_\theta(\{\mathbf{h}^{(l)}\}_{l=0}^{L})$. The LLM remains frozen and we train only $\theta$, enabling retrofitting to deployed models while reusing the serving forward pass. The main challenge is to aggregate information across both tokens ($T$) and layers ($L$) to extract discriminative features for classification.

\subsection{Two-Stage Aggregation Architecture}

Our architecture addresses this dimensionality challenge with a two-stage aggregation scheme. The LLM produces a three-dimensional hidden-state tensor of shape $L \times T \times d$ (layers $\times$ tokens $\times$ hidden dimension). To avoid feeding this tensor directly to a lightweight classifier, we first aggregate across tokens within each layer to obtain $L+1$ fixed-size layer summaries, then aggregate across layers to produce a single $d$-dimensional vector that is consumed by a standard classification head.

\paragraph{Overview.}
Figure~\ref{fig:architecture} summarizes our pipeline. Given hidden states $\{\mathbf{h}^{(l)}\}_{l=0}^{L}$, we (Stage~1) aggregate tokens within each layer to obtain layer summaries $\mathbf{v}^{(l)} \in \mathbb{R}^d$, (Stage~2) aggregate across layers to produce a single vector $\mathbf{v} \in \mathbb{R}^d$, and then apply a linear head to obtain class logits. We use the same aggregation mechanism for both stages (Section~\ref{ssec:aggregation-mechanisms}).

\paragraph{Stage 1: Token-level aggregation.}
For each layer $l$, we reduce $\mathbf{h}^{(l)} \in \mathbb{R}^{T \times d}$ to a layer summary using $\mathcal{A}^{(l)}_{\text{token}}$:
\[
\mathbf{v}^{(l)} = \mathcal{A}^{(l)}_{\text{token}}(\mathbf{h}^{(l)}), \qquad \mathbf{v}^{(l)} \in \mathbb{R}^{d},
\]
yielding $\{\mathbf{v}^{(l)}\}_{l=0}^L$.

\paragraph{Stage 2: Layer-level aggregation.}
We then aggregate across depth with $\mathcal{A}_{\text{layer}}$:
\[
\mathbf{v} = \mathcal{A}_{\text{layer}}(\{\mathbf{v}^{(l)}\}_{l=0}^L), \qquad \mathbf{v} \in \mathbb{R}^{d}.
\]
This module learns task-specific layer weighting from data, avoiding manual layer selection.

\paragraph{Classification head.}
Finally, logits are produced by a linear head,
\[
\text{logits} = \mathbf{W}_{\text{out}}\mathbf{v} + \mathbf{b}_{\text{out}}, \qquad \mathbf{W}_{\text{out}} \in \mathbb{R}^{C \times d},
\]
trained with cross-entropy. At inference time, the probe emits a label
from the same forward pass that initiates generation. If the input is
classified as unsafe, the orchestration layer can halt generation before
any tokens are streamed and either return a templated refusal or
re-prompt the serving LLM to produce a contextual rejection, with no
additional model invocation.

\subsection{Aggregation Mechanisms}\label{ssec:aggregation-mechanisms}

The core design choice is the aggregation operator that compresses representations across tokens (Stage~1) and across layers (Stage~2). This operator must balance expressiveness with overhead: it should extract task-relevant signal while adding minimal parameters and inference cost.

We study three mechanisms spanning a simple--to--expressive spectrum, from fixed pooling to learned attention. Here, we use the same mechanism for Stage~1 and Stage~2 to keep the architecture uniform and to enable controlled comparisons.

Let $\mathbf{X} \in \mathbb{R}^{N \times d}$ denote the input to an aggregation block, where $N \in \{T, L+1\}$ depends on the stage. The block outputs a fixed-size vector $\mathbf{v} \in \mathbb{R}^d$ summarizing the most informative content in $\mathbf{X}$.

\paragraph{Direct pooling.}
The simplest aggregation strategy. It applies a fixed pooling operator over valid (non-padding) positions. Max pooling selects the largest activation per dimension,
\[
\mathbf{v}[j] = \max_{i \in \mathcal{V}} \mathbf{X}[i, j],
\]
while mean pooling averages activations,
\[
\mathbf{v}[j] = \frac{1}{|\mathcal{V}|}\sum_{i \in \mathcal{V}} \mathbf{X}[i, j],
\]
where $\mathcal{V}$ denotes valid positions.

\paragraph{Scoring attention gate.}\label{sssec:scoring-attention-gate}
To add learnable weighting with minimal parameters, this approach learns to assign scalar importance scores to each input (token or layer) using a single linear projection,
\[
s_i = \tanh(\mathbf{w}^\top \mathbf{X}[i,:]), \qquad i \in \{1,\dots,N\},
\]
mask padding by setting $s_i=-\infty$, and normalize with softmax:
\[
\boldsymbol{\alpha}=\mathrm{softmax}(\mathbf{s}), \qquad 
\mathbf{v}=\sum_{i=1}^{N}\alpha_i\,\mathbf{X}[i,:].
\]
For token-level aggregation, we use one gate per layer ($L+1$ total); for layer-level aggregation, we use a single shared gate.
\textbf{Parameter count:} $(L{+}2)d$ trainable parameters.

\paragraph{Multi-head self-attention.}
We use multi-head self-attention~\cite{NIPS2017_3f5ee243} as a more expressive alternative. We implement attention with PyTorch \texttt{scaled\_dot\_product\_attention}, which can leverage FlashAttention when available~\cite{dao2023flashattention2fasterattentionbetter}. To control cost, we downcast QKV projections from $d$ to $d_{\text{inner}}<d$ (e.g., $d/16$ or $d/32$), split into $H$ heads with $d_{\text{head}}=d_{\text{inner}}/H$. For each head $h$, we compute
\[
\text{Attn}_h(\mathbf{Q}_h, \mathbf{K}_h, \mathbf{V}_h) = \text{softmax}\left(\frac{\mathbf{Q}_h\mathbf{K}_h^\top}{\sqrt{d_{\text{head}}}}\right)\mathbf{V}_h,
\]
concatenate heads, project back to $d$ dimensions, which we then pool (mean or max) over the sequence dimension to obtain $\mathbf{v} \in \mathbb{R}^d$.

As with the scoring gate, we use $L{+}2$ MHA modules (one per layer for Stage~1, one for Stage~2), for a total of $(L{+}2)\cdot 4d\,d_{\text{inner}}$ parameters.

\section{Experiments and Results}\label{sec:experiments}

\newcommand{\yes}{\textcolor{black}{\checkmark}}
\newcommand{\no}{--}

\begin{table*}[t]
\centering
\small
\begin{tabular}{lcccc}
\toprule
\textbf{Method} & \textbf{F1(\%)} & \textbf{AUPRC} & \textbf{Added Params (M)} & \textbf{Extra LM call} \\
\midrule
\multicolumn{5}{l}{\textbf{Standalone classifiers (extra model call)}}\\
ToxicChat-T5-large$^{\dagger}$ & 82.2 & 0.885 & 780 & \checkmark \\
\addlinespace
\multicolumn{5}{l}{\textbf{Reuse baselines (same serving pass)}}\\
MULI (logits)$^{\dagger}$ & 77.8 & 0.829 & 0.13 & -- \\
\addlinespace
\multicolumn{5}{l}{\textbf{Ours (same serving pass)}}\\
\multicolumn{5}{l}{\textit{Llama-3.2-3B backbone}}\\
\quad Direct pooling & 73.53 $\pm$ 0.68 & 0.812 $\pm$ 0.005 & 0.003 & -- \\
\quad Scoring attention & 80.49 $\pm$ 1.17 & 0.854 $\pm$ 0.008 & 0.10 & -- \\
\quad Multi-head self-attn & \underline{84.51 $\pm$ 0.43} & 0.898 $\pm$ 0.005 & 35 & -- \\
\addlinespace
\multicolumn{5}{l}{\textit{GPT-OSS-20B backbone}}\\
\quad Direct pooling & 77.36 $\pm$ 0.87 & 0.833 $\pm$ 0.008 & 0.003 & -- \\
\quad Scoring attention & 79.23 $\pm$ 0.58 & 0.854 $\pm$ 0.005 & 0.08 & -- \\
\quad Multi-head self-attn & \textbf{86.17 $\pm$ 0.51} & \textbf{0.915 $\pm$ 0.004} & 27 & -- \\
\addlinespace
\multicolumn{5}{l}{\textit{Qwen3-30B-A3B backbone}}\\
\quad Direct pooling & 73.24 $\pm$ 0.52 & 0.827 $\pm$ 0.005 & 0.002 & -- \\
\quad Scoring attention & 80.94 $\pm$ 0.62 & 0.883 $\pm$ 0.006 & 0.10 & -- \\
\quad Multi-head self-attn & 83.76 $\pm$ 0.9 & \underline{0.905 $\pm$ 0.005} & 26 & -- \\
\bottomrule
\end{tabular}
\caption{\textbf{ToxicChat (in-distribution).} All \textit{Reuse} methods (MULI and ours) attach a probe to a frozen LLM backbone and reuse the same forward pass; we therefore report \textbf{added parameters beyond the backbone}. Standalone baselines require an \textbf{additional model invocation}. $\dagger$From~\cite{lin-etal-2023-toxicchat}.}
\label{tab:toxichat-indist}
\end{table*}

\begin{table*}[t]
\centering
\small
\begin{tabular}{lcccc}
\toprule
\textbf{Method} & \textbf{F1(\%)} & \textbf{AUPRC} & \textbf{Added Params (M)} & \textbf{Extra LM call} \\
\midrule

\multicolumn{5}{l}{\textbf{Standalone guard models / APIs (extra model call)}}\\
OpenAI Moderation$^{\dagger}$ & 61.4 & 0.631 & --- & \checkmark \\
Llama Guard$^{\ddagger,\S}$ & 61.6 & 0.626 & 7{,}000 & \checkmark \\
Llama-Guard2$^{\ddagger}$ & 47.1 & --- & 8{,}000 & \checkmark \\
Aegis-Guard-D$^{\ddagger}$ & 70.0 & --- & 8{,}000 & \checkmark \\
GPT-4$^{\ddagger}$ & 68.3 & --- & 1{,}700{,}000 & \checkmark \\
WildGuard$^{\ddagger}$ & \underline{70.8} & --- & 7{,}000 & \checkmark \\

\addlinespace
\multicolumn{5}{l}{\textbf{Reuse baselines (same serving pass)}}\\
ShieldHead (Llama3.1-8B)$^{\|}$ & 64.3 & --- & 91 & -- \\
ShieldHead (Gemma2-27B)$^{\|}$ & 67.7 & --- & 306 & -- \\

\addlinespace
\multicolumn{5}{l}{\textbf{Ours (same serving pass; trained on WildGuardMix)}}\\
Direct pooling & 53.33 & 0.565 & 0.003 & -- \\
Scoring attention & 64.81 & \underline{0.706} & 0.10 & -- \\
Multi-head self-attn & \textbf{72.88} & \textbf{0.798} & 35 & -- \\
\bottomrule
\end{tabular}
\caption{\textbf{ToxicChat (out-of-distribution).} Guard models / APIs require an \textbf{additional model invocation}. Our probes attach to a \textbf{frozen Llama-3.2-3B} serving model and reuse the same forward pass; we therefore report \textbf{added parameters beyond the 3B backbone}. $\dagger$From~\cite{lin-etal-2023-toxicchat}. $\ddagger$From~\cite{han2024wildguardopenonestopmoderation}. $\S$AUPRC from~\cite{inan2023llamaguardllmbasedinputoutput}. $\|$From~\cite{xuan-etal-2025-shieldhead}.}
\label{tab:toxichat-ood}
\end{table*}

\begin{table*}[t]
\centering
\small
\begin{tabular}{lccc}
\toprule
\textbf{Method} & \textbf{F1(\%)} & \textbf{Added Params (M)} & \textbf{Extra LM call} \\
\midrule

\multicolumn{4}{l}{\textbf{Standalone guard models / APIs (extra model call)}}\\
OpenAI Mod API & 12.1 & --- & \checkmark \\
Llama-Guard2 & 70.4 & 8{,}000 & \checkmark \\
Llama3.1-AegisGuard & 82.1 & 8{,}000 & \checkmark \\
WildGuard & \textbf{88.9} & 7{,}000 & \checkmark \\
\addlinespace

\multicolumn{4}{l}{\textbf{Reuse baselines (same serving pass)}}\\
MULI (logits, Llama-3.2-3B) & 83.79 & 0.13 & -- \\
\addlinespace

\multicolumn{4}{l}{\textbf{Ours (same serving pass)}}\\
Direct pooling & 82.84 $\pm$ 0.06 & 0.003 & -- \\
Scoring attention & 85.98 $\pm$ 0.61 & 0.10 & -- \\
Multi-head self-attn & \underline{88.55 $\pm$ 0.30} & 35 & -- \\
\bottomrule
\end{tabular}
\caption{\textbf{WildGuardMix.} \textit{Reuse} methods attach a probe to a \textbf{serving Llama-3.2-3B} model and reuse the forward pass; we report \textbf{added parameters beyond the backbone}. Standalone baselines require an \textbf{additional model invocation}.}
\label{tab:wildguard}
\end{table*}

\begin{table*}[t]
\centering
\small
\begin{tabular}{lccccc}
\toprule
\textbf{Method} & \textbf{IMDB} & \textbf{SST-2} & \textbf{Emotion} &
\textbf{Added Params (M)} & \textbf{Extra LM call} \\
\midrule

\multicolumn{6}{l}{\textbf{Standalone classifiers / truncation baselines (extra model call)}}\\
DeBERTa V3 Large & \underline{95.34} & 90.38 & \underline{87.65} & 418 & \checkmark \\
RoBERTa Large & 94.30 & \textbf{95.99} & 84.16 & 355 & \checkmark \\
SentriLlama 3.2 (3B) Instruct$^{\ddagger}$ & \best{95.79} & \underline{95.94} & 82.20 & 0.003--1.2 & \checkmark \\
\addlinespace

\multicolumn{6}{l}{\textbf{Reuse baselines (same serving pass)}}\\
MULI (logits, Llama-3.2-3B) & 86.50 & 93.19 & 64.05 & 0.13--0.77 & -- \\
\addlinespace

\multicolumn{6}{l}{\textbf{Prompting (extra model call)}}\\
Llama 3.2 (3B) Zero-shot & 77.59 & 83.97 & 44.55 & 0 & \checkmark \\
Llama 3.2 (3B) Few-shot & 76.06 & 85.28 & 33.40 & 0 & \checkmark \\
Llama 3.2 (3B) Chain-of-Thought & 91.54 & 93.06 & 56.05 & 0 & \checkmark \\
\addlinespace

\multicolumn{6}{l}{\textbf{Ours (same serving pass)}}\\
Direct pooling & 94.07 $\pm$ 0.04 & 92.29 $\pm$ 0.18 & 69.18 $\pm$ 0.64 & 0.003--0.018 & -- \\
Scoring attention & 95.05 $\pm$ 0.11 & 94.42 $\pm$ 0.64 & 84.58 $\pm$ 0.67 & 0.10--0.11 & -- \\
Multi-head self-attn & 95.15 $\pm$ 0.06 & 95.39 $\pm$ 0.32 & \best{87.68 $\pm$ 1.09} & 35--35.5 & -- \\
\bottomrule
\end{tabular}
\caption{Sentiment and emotion classification on IMDB, SST-2, and Emotion. For \textit{Reuse} methods (MULI and ours), we report \textbf{added parameters beyond a frozen Llama-3.2-3B} serving model and no extra model call at inference. Baseline and prompting results from~\citet{dipalma2025llamasfeelingstoounveiling}. $\ddagger$SentriLlama truncates the backbone at a selected layer and trains a classifier head; the reported trainable range varies by classifier and number of classes.}
\label{tab:sentiment_results}
\end{table*}

\subsection{Experimental Setup}

\noindent\textbf{Model.} Our primary serving backbone is \texttt{Llama-3.2-3B-Instruct}~\citep{grattafiori2024llama3herdmodels}, used across all benchmarks.\footnote{Code available at \url{https://github.com/gmeyoyan/LLM-Probing-Classifiers}} To validate generalization beyond a single model family, we additionally evaluate on two mixture-of-experts architectures on ToxicChat: \texttt{GPT-OSS-20B (20B total / 3.6B active)} \cite{openai2025gptoss120bgptoss20bmodel} and \texttt{Qwen3-30B-A3B (30B total / 3B active)} \cite{yang2025qwen3technicalreport}. Together, these three backbones span dense and sparse-routing designs across a 10$\times$ total parameter range. Notably, even the compact Llama-3 3B dense backbone already exposes hidden representations that support strong moderation and NLU probes. Since our probes read out from the serving model's internal representations without modifying the backbone, stronger backbones offer a natural path to improved probe accuracy at essentially unchanged \emph{relative} probe overhead; established scaling laws support this expectation~\citep{kaplan2020scaling,hoffmann2022training}. Our multi-backbone results (Table~\ref{tab:toxichat-indist}) confirm that the approach generalizes across architectures: the method ranking and competitive performance hold across all three models despite fundamentally different routing mechanisms.

\noindent\textbf{Datasets.}
We evaluate on five public benchmarks spanning two settings: safety moderation and sentiment classification.

\noindent\textbf{ToxicChat.} ToxicChat~\citep{lin-etal-2023-toxicchat} contains 10,166 human--LLM prompts \citep{zheng2023judging} labeled for toxicity, including explicit toxic content and jailbreak-style prompts. We use the authors' train/test split and restrict the test set to human-annotated samples.

\noindent\textbf{WildGuardMix.} WildGuardMix~\citep{han2024wildguardopenonestopmoderation} is a large safety benchmark ($\sim$ 89K examples) mixing multiple sources and synthetic data across diverse harm categories (e.g., violence, hate, self-harm, sexual content, privacy), with both direct requests and adversarial jailbreaks. We use the standard train/test split.

\noindent\textbf{IMDB / SST-2 / Emotion.}
IMDB~\citep{maas2011learning} is binary sentiment over movie reviews; following~\citet{dipalma2025llamasfeelingstoounveiling}, we use their 12,500-example subset (6,250/6,250). SST-2~\citep{socher2013recursive} is binary sentiment over short snippets; following~\citet{dipalma2025llamasfeelingstoounveiling}, we use their 7,000-example subset. Emotion~\citep{saravia-etal-2018-carer} is a multi-class emotion classification benchmark; following~\citet{dipalma2025llamasfeelingstoounveiling}, we use their 8,000-example subset.

\noindent\textbf{Metrics.}
We report F1 for ToxicChat and WildGuardMix, and accuracy for IMDB, SST-2, and Emotion; for ToxicChat we additionally report AUPRC when available.

\subsection{Hyperparameter Search}

A practical benefit of our probes is their small trainable footprint, which makes extensive hyperparameter exploration feasible. We explored approximately 100 configurations per dataset on the validation set, a scale of experimentation that would be prohibitive when fine-tuning multi-billion parameter guard models. This allows us to select optimal configurations and analyze how sensitive results are to different hyperparameter values. We present such analysis in Appendix~\ref{app:hyperparameter-search}.

To reduce peak GPU memory during training, we optionally precompute and cache hidden states from the frozen LLM before training the probe. This decouples backbone inference from classifier training, allowing larger batch sizes under VRAM constraints. At inference time, the pipeline is unchanged: a single forward pass of the serving LLM with the probe attached.

For each dataset and aggregation mechanism, we search standard training hyperparameters (learning rate, batch size, weight decay) as well as mechanism-specific choices. For MHA, we vary the number of heads and the downcasting factor; for pooling and MHA, we also vary the pooling operator (mean vs.\ max). Full ranges and selected settings are reported in Appendix~\ref{app:hyperparameter-search}.

\subsection{Safety Classification Results}

We evaluate whether safety signals in a frozen serving LLM can be read out with lightweight probes that run in the same forward pass as generation. To make the deployment trade-off explicit, we group baselines by whether they require an extra LM call at inference (standalone guard models / APIs) versus reuse methods that attach a small head to the serving model (MULI and ours). For all reuse methods, we report added parameters beyond those from the frozen backbone.

\noindent\textbf{ToxicChat: in-distribution (Table~\ref{tab:toxichat-indist}).}
When trained on the ToxicChat training set and evaluated on the held-out test set (restricted to human-annotated samples), learned token--layer aggregation improves over fixed readouts. Direct pooling reaches 73.53 F1, while the scoring-attention gate improves to 80.49 F1 with 0.10M added parameters. The multi-head self-attention (MHA) probe achieves 84.51 F1 and 0.898 AUPRC, improving over the logit-reuse baseline MULI (77.8 F1 / 0.829 AUPRC) and the standalone ToxicChat-T5 classifier (82.2 F1 / 0.885 AUPRC), while avoiding an extra model invocation. These gains are obtained with aggressive attention downcasting, suggesting that the improvement is driven primarily by where we read out (across tokens and layers) rather than by large additional capacity.

\noindent\textbf{Cross-backbone consistency (Table~\ref{tab:toxichat-indist}).}
To test whether these findings depend on a specific model family, we repeat the ToxicChat evaluation on two mixture-of-experts backbones: GPT-OSS-20B and Qwen3-30B-A3B. The method ranking (pooling $<$ scoring-attention $<$ MHA) is preserved across all three architectures. All MHA probes match or exceed the standalone ToxicChat-T5-large baseline (82.2 F1) without an extra model call, with GPT-OSS-20B achieving the highest F1 (86.17) and AUPRC (0.915). These results suggest that the layer-distributed safety signal is neither an artifact of the Llama architecture nor of dense transformer structure, as MoE models (which route tokens through sparse expert subsets) exhibit the same pattern.

\noindent\textbf{ToxicChat: cross-dataset generalization (Table~\ref{tab:toxichat-ood}).}
To test robustness under distribution shift, we train on WildGuardMix and evaluate on ToxicChat. Pooling drops to 53.33 F1, indicating that learned selection is important for transfer. The scoring-attention gate reaches 64.81 F1 / 0.706 AUPRC with 0.10M parameters, and MHA reaches 72.88 F1 / 0.798 AUPRC. In this setting, our probes outperform several guard-model and API baselines reported in Table~\ref{tab:toxichat-ood} while requiring no additional model call at inference.
Because these baselines vary in architecture and training data, we treat such cross-dataset comparisons as indicative of the compute--accuracy trade-off rather than controlled head-to-head matches. The key result is that probes trained on a broad safety mixture transfer to ToxicChat without deploying a separate guard LLM.

\noindent\textbf{WildGuardMix (Table~\ref{tab:wildguard}).}
On the larger and more heterogeneous WildGuardMix benchmark, simple reuse is strong: direct pooling achieves 82.84 F1, and learned aggregation provides consistent gains. The scoring-attention gate reaches 85.98 F1 with 0.10M added parameters, and MHA achieves 88.55 F1, approaching the strongest standalone guard baseline in the table (WildGuard at 88.9 F1) while training only 35M parameters and still running in the same serving pass. Compared to MULI (83.79 F1), which reads out from first-token logits, both learned aggregation variants perform better, supporting the hypothesis that safety cues are distributed across layers and token positions and are better captured by joint token--layer selection.

\noindent\textbf{Takeaway.}
Across both safety datasets, method ranking is consistent (pooling $<$ scoring-attention $<$ MHA), and multi-backbone experiments on ToxicChat confirm this ordering generalizes across dense and MoE architectures. Overall, the benefit is not only reuse of the serving computation, but learning where in the $L \times T \times d$ tensor the safety signal is most separable, reducing the latency and VRAM overhead associated with a second LLM in the orchestration pipeline.

\begin{figure*}[t]
    \centering
    \includegraphics[width=1\linewidth]{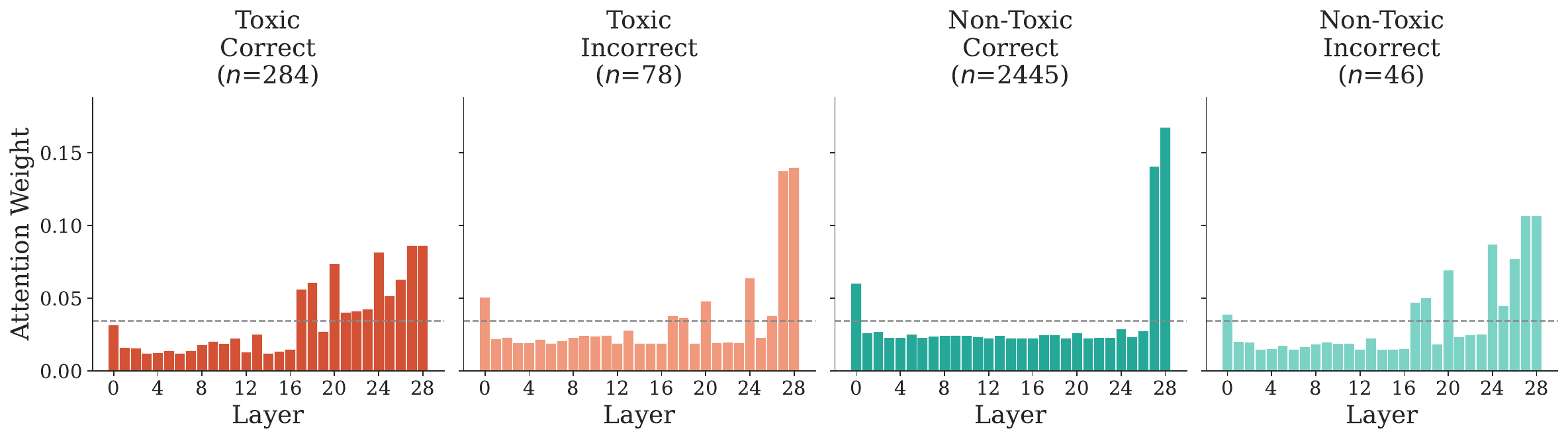}
\caption{Attention weights from the scoring-attention probe on ToxicChat, stratified by label and correctness. The dashed line marks uniform $1/(L+1)$ ($\approx 0.034$). Toxic prompts attend to later layers (L17--L28), while non-toxic prompts concentrate on layers L0 and L27--L28. Misclassifications resemble the predicted-class profile.}

    \label{fig:layer-attention-tc}
\end{figure*}

\subsection{Sentiment Analysis Results}

Table~\ref{tab:sentiment_results} reports accuracy on IMDB, SST-2, and Emotion. Overall, we find that a frozen \texttt{Llama-3.2-3B-Instruct} already encodes strong sentiment/emotion cues in its hidden-state tensor, and that a lightweight readout trained on these representations substantially outperforms prompting while avoiding an extra model invocation.

\noindent\textbf{Reuse vs.\ prompting.}
Prompting the same backbone is markedly weaker: zero-shot and few-shot prompting perform poorly on all three datasets, and even chain-of-thought remains far below trained probes. In contrast, our reuse probes deliver high accuracy with a single serving pass, showing that supervised readouts can reliably extract sentiment information from the internal representations without changing the backbone.

\noindent\textbf{Effect of learned token--layer aggregation.}
Within our probe family, learned aggregation consistently wins over pooling. Direct pooling is already strong on IMDB and SST-2 but underperforms significantly on Emotion (69.18), suggesting that simple summaries may miss task-relevant structure. The scoring-attention gate provides a large jump with minimal additional capacity, using only 0.10--0.11M added parameters, indicating that learning where to read across tokens and layers captures much of the signal. The more expressive MHA probe yields the best overall reuse performance, and is particularly beneficial for the multi-class Emotion task.

\noindent\textbf{Comparison to standalone classifiers and logit reuse.}
Against standalone classifiers (DeBERTa, RoBERTa), our MHA probe is competitive on IMDB and SST-2 and achieves the best result on Emotion. Compared to the logit-reuse baseline MULI, which reads out from first-token logits, our probes are stronger, especially on Emotion (87.68 vs.\ 64.05), suggesting that hidden states and token--layer aggregation provide a richer signal than logits alone.

\noindent\textbf{Relation to SentriLlama.}
SentriLlama reports strong results by truncating the backbone at a selected layer and training a classifier head. In contrast, our probes operate on an intact frozen serving model and can run alongside generation, which is the common requirement in orchestration settings. Despite this constraint, our scoring-attention and MHA probes remain competitive across datasets while keeping inference to a single serving pass.

\noindent\textbf{Takeaway.}
For recurring sentiment and emotion classification inside an LLM orchestration, reuse probes offer a favorable trade-off: large gains over prompting, improvements over logit-only reuse, and accuracy competitive with task-specific classifiers while preserving a single-pass serving pipeline.

\section{Layer Attention Analysis}
\label{sec:attention-analysis}

A central motivation of our approach is that task-relevant information is distributed across depth, and that committing to a fixed readout position may underutilize available signal. To probe this, we visualize the layer-aggregation weights produced by the scoring-attention gate on ToxicChat, stratified by label and prediction correctness (Figure~\ref{fig:layer-attention-tc}).

No single layer dominates across all groups. For classified toxic prompts, attention is spread over later layers (L17--L28), with layers receiving weight above the uniform baseline. This pattern is consistent with features emerging progressively through depth, suggesting that restricting the readout to a single layer can discard evidence from nearby layers. In contrast, correctly classified non-toxic prompts place most mass on the final layers (L27--L28), with a smaller contribution from the embedding layer (L0).

These class-conditional patterns can be interpreted as the probe emphasizing different depths where representations are most separable for each content type. Toxic prompts tend to elicit informative intermediate-to-late representations starting around L17, whereas non-toxic prompts rely more heavily on the final layers.

Regarding misclassifications, toxic prompts predicted as non-toxic concentrate attention on L28, resembling the correctly classified non-toxic profile; conversely, misclassified non-toxic prompts exhibit the more distributed pattern typical of toxic predictions. Thus, errors are associated with layer-weight profiles that align more with the predicted class than with the ground-truth label, which may reflect atypical representations, label noise, or genuinely ambiguous cases.

These findings align with classical BERTology work showing that transformer layers encode different abstractions \citep{devries2020whats,Jawahar2019}, and extend it to decoder-only LLMs: safety-relevant features are not localized to a single depth but distributed across the network, motivating learned aggregation over fixed readouts. Similar layer-wise attention patterns emerge 
for sentiment classification (Figure \ref{fig:layer-attention-sst2} in Appendix \ref{app:attention-analysis}).

\section{Inference Efficiency Analysis}
\label{sec:efficiency}

\begin{table}[t!]
\centering
\scriptsize
\setlength{\tabcolsep}{4pt}
\begin{tabular}{lrrr}
\toprule
\textbf{Approach} & \textbf{Throughput} & \textbf{Avg. Latency} & \textbf{Peak GPU} \\
 & \textbf{(samples/s)} & \textbf{(ms/sample)} & \textbf{(MB)} \\
\midrule
Base (3B) & 37.84 & 26.43 & 6497.63 \\
+ Pooling probe & 33.72 & 29.66 & 6497.63 \\
+ Scoring probe & 32.36 & 30.90 & 6749.00 \\
+ MHA probe & 24.83 & 40.27 & 6968.92 \\
+ ToxicChat-T5 (780M) & 11.32 & 88.33 & 7992 \\
+ WildGuard (7B) & 9.97 & 100.32 & 21084 \\
+ Llama Guard 3 (8B) & 8.12 & 123.21 & 22769 \\
\bottomrule
\end{tabular}
\caption{Latency benchmark on ToxicChat test prompts using Llama-3.2-3B-Instruct as the serving model (2000 samples, max length 512, batch size 1, \texttt{max\_new\_tokens=1}). Probe rows reuse the serving forward pass. Guard-model rows (ToxicChat-T5, WildGuard, Llama Guard~3) run a separate model followed by the serving LLM, covering the architectural families behind the majority of open-source guard models presented in our previous safety benchmarks. }
\label{tab:latency}
\end{table}

We quantify the systems overhead of our probes relative to a standard production pattern that runs a separate guard model before invoking the serving LLM. We report end-to-end latency, throughput, and peak GPU memory, since these directly determine serving cost and tail-latency budgets.

\noindent\textbf{Deployment scenarios.}
Our baseline is a \emph{guard-then-serve} pipeline in which a separate guard model screens prompts before the serving \texttt{Llama-3.2-3B-Instruct} is invoked. We benchmark three representative guard models: ToxicChat-T5 (780M), WildGuard (7B, Mistral fine-tune), and Llama Guard~3 (8B)~\citep{grattafiori2024llama3herdmodels}. In contrast, our probes are attached to the serving model and compute the label from hidden states produced during the same forward pass, eliminating an extra model invocation.

\noindent\textbf{Results.}
Table~\ref{tab:latency} shows a clear separation between single-pass reuse and guard-then-serve pipelines. Even against the smallest standalone baseline (ToxicChat-T5, 780M), the MHA probe is over 2$\times$ faster and uses 1\,GB less peak memory. Relative to the serving-only baseline, pooling adds minimal overhead and the scoring-attention gate remains close. The MHA probe is slower, reflecting the additional attention computation, but still substantially faster than any guard-then-serve configuration. Throughput follows the same trend.

Peak memory is dominated by whether a second model is loaded. All probe variants remain near the serving footprint (6.5--7.0\,GB), whereas guard-then-serve pipelines range from 8.0\,GB (ToxicChat-T5) to 22.8\,GB (Llama Guard~3) depending on guard model size. Overall, these measurements show that (i) reuse-based classification preserves a single-model serving profile, and (ii) within our probe family, pooling and scoring-attention offer the best latency--accuracy trade-off, while MHA provides higher accuracy at a moderate but still single-pass overhead.

\section{Conclusions and Future Work}

We showed that lightweight probes trained on a frozen serving LLM’s hidden states can support moderation and sentiment/emotion classification without deploying a separate classifier model. By casting classification as representation selection over the full $L \times T \times d$ hidden-state tensor and using a two-stage token--layer aggregation, our approach avoids committing to a single layer or token while remaining parameter-efficient. Across ToxicChat and WildGuardMix, our probes approach or match strong guard-model baselines with far fewer trainable parameters, and they are competitive on IMDB, SST-2, and Emotion. Multi-backbone experiments on dense and mixture-of-experts architectures confirm that the method ranking and competitive performance are not specific to a single model family. System-wise, reuse enables a single-pass pipeline that reduces latency, VRAM, and orchestration
complexity relative to guard-then-serve deployments. Key next steps include extending multi-backbone evaluation to additional tasks, testing with larger active parameter counts, and exploring fine-grained safety taxonomies.

\section*{Limitations}

\paragraph{Generalization Across Models and Architectures.} We validated cross-architecture robustness on ToxicChat using three backbone families (Llama-3.2-3B, GPT-OSS-20B, Qwen3-30B-A3B), but sentiment and emotion benchmarks were evaluated only on
Llama-3.2-3B-Instruct. Moreover, all three backbones share approximately 3B active parameters, so the effect of substantially larger active capacity on probe accuracy remains untested. Different architectures may encode task-relevant information differently across layers, potentially requiring architecture-specific tuning of our aggregation mechanisms.

\paragraph{Sequence Length Constraints.} Our experimental setup was limited by VRAM constraints when processing longer sequences with larger batch sizes. Longer input sequences (e.g., full documents, multi-turn conversations) may present memory challenges that require optimization techniques.

\paragraph{Training Data Requirements.} Our smallest dataset (SST-2) contained 7,000 examples. The minimum dataset size required to effectively train these probes remains unclear, particularly for the more parameter-intensive multi-head attention variants. Tasks with limited labeled data may require investigation of few-shot or transfer learning approaches.

\paragraph{Response Generation Limitation.} Unlike safety fine-tuning or guard models that can generate explanatory rejections (e.g., ``I cannot help with that request because...''), our classification approach can only detect harmful content and interrupt generation. 
However, because our probes complement rather than replace the serving LLM, the orchestration layer can trigger a refusal response from the same model once unsafe content is detected; the serving model's generation capability is available by design. We did not evaluate such conditional re-prompting strategies, and integrating probe-based detection with contextual refusal generation is a natural direction for future work. 

\paragraph{Robustness and Failure Scenarios.} Our OOD results (Table~\ref{tab:toxichat-ood}) show that distribution shift degrades probe performance. Because probes are lightweight classifiers over learned feature patterns, they may fail silently on novel harm categories, evolving jailbreak strategies, or input distributions not represented in training data. Unlike full guard LLMs, which can reason over text and generalize more broadly, probes cannot adapt without retraining. In safety-critical deployments with rapidly evolving threat landscapes, or in domains requiring coverage of harm categories beyond the training distribution, a dedicated guard model may remain necessary despite its higher computational cost.

\section*{Ethical Considerations}

\paragraph{Misuse potential.}
Understanding where safety signals are encoded in LLM hidden states could inform adversarial attacks designed to evade detection. However, similar information is available in prior interpretability work, and we believe the benefits of efficient moderation outweigh this risk.

\paragraph{Bias and disparate impact.}
Because probes read out from a frozen backbone, they inherit biases and representation gaps from the underlying LLM and training data. This can lead to disparate error rates across demographic groups.

\paragraph{Privacy and data handling.}
Our experiments optionally cache hidden states to reduce training-time GPU memory. Hidden representations may encode sensitive information present in prompts. Cached activations should therefore be treated as sensitive data, with appropriate access controls and limited retention.

\bibliography{custom}

\begin{thebibliography}{27}
\providecommand{\natexlab}[1]{#1}

\bibitem[{Chrabąszcz et~al.(2025)Chrabąszcz, Szatkowski, Wójcik, Dubiński, Trzciński, and Cygert}]{chrabąszcz2025llmsunderstandsafetyinputs}
Maciej Chrabąszcz, Filip Szatkowski, Bartosz Wójcik, Jan Dubiński, Tomasz Trzciński, and Sebastian Cygert. 2025.
\newblock \href {https://arxiv.org/abs/2502.16174} {Do llms understand the safety of their inputs? training-free moderation via latent prototypes}.
\newblock \emph{Preprint}, arXiv:2502.16174.

\bibitem[{Dao(2023)}]{dao2023flashattention2fasterattentionbetter}
Tri Dao. 2023.
\newblock \href {https://arxiv.org/abs/2307.08691} {Flashattention-2: Faster attention with better parallelism and work partitioning}.
\newblock \emph{Preprint}, arXiv:2307.08691.

\bibitem[{de~Vries et~al.(2020)de~Vries, van Cranenburgh, and Nissim}]{devries2020whats}
Wietse de~Vries, Andreas van Cranenburgh, and Malvina Nissim. 2020.
\newblock \href {https://doi.org/10.18653/v1/2020.findings-emnlp.389} {What{'}s so special about {BERT}{'}s layers? a closer look at the {NLP} pipeline in monolingual and multilingual models}.
\newblock In \emph{Findings of the Association for Computational Linguistics: EMNLP 2020}, pages 4339--4350, Online. Association for Computational Linguistics.

\bibitem[{Ghosh et~al.(2024)Ghosh, Varshney, Galinkin, and Parisien}]{ghosh2024aegisonlineadaptiveai}
Shaona Ghosh, Prasoon Varshney, Erick Galinkin, and Christopher Parisien. 2024.
\newblock \href {https://arxiv.org/abs/2404.05993} {Aegis: Online adaptive ai content safety moderation with ensemble of llm experts}.
\newblock \emph{Preprint}, arXiv:2404.05993.

\bibitem[{Grattafiori et~al.(2024)Grattafiori, Dubey, Jauhri, Pandey, Kadian, Al-Dahle, Letman, Mathur, Schelten, Vaughan, Yang, Fan, Goyal, Hartshorn, Yang, Mitra, Sravankumar, Korenev, Hinsvark, Rao, Zhang, Rodriguez, Gregerson, Spataru, Roziere, Biron, Tang, Chern, Caucheteux, Nayak, Bi, Marra, McConnell, Keller, Touret, Wu, Wong, Ferrer, Nikolaidis, Allonsius, Song, Pintz, Livshits, Wyatt, Esiobu, Choudhary, Mahajan, Garcia-Olano, Perino, Hupkes, Lakomkin, AlBadawy, Lobanova, Dinan, Smith, Radenovic, Guzmán, Zhang, Synnaeve, Lee, Anderson, Thattai, Nail, Mialon, Pang, Cucurell, Nguyen, Korevaar, Xu, Touvron, Zarov, Ibarra, Kloumann, Misra, Evtimov, Zhang, Copet, Lee, Geffert, Vranes, Park, Mahadeokar, Shah, van~der Linde, Billock, Hong, Lee, Fu, Chi, Huang, Liu, Wang, Yu, Bitton, Spisak, Park, Rocca, Johnstun, Saxe, Jia, Alwala, Prasad, Upasani, Plawiak, Li, Heafield, Stone, El-Arini, Iyer, Malik, Chiu, Bhalla, Lakhotia, Rantala-Yeary, van~der Maaten, Chen, Tan, Jenkins, Martin, Madaan, Malo, Blecher,
  Landzaat, de~Oliveira, Muzzi, Pasupuleti, Singh, Paluri, Kardas, Tsimpoukelli, Oldham, Rita, Pavlova, Kambadur, Lewis, Si, Singh, Hassan, Goyal, Torabi, Bashlykov, Bogoychev, Chatterji, Zhang, Duchenne, Çelebi, Alrassy, Zhang, Li, Vasic, Weng, Bhargava, Dubal, Krishnan, Koura, Xu, He, Dong, Srinivasan, Ganapathy, Calderer, Cabral, Stojnic, Raileanu, Maheswari, Girdhar, Patel, Sauvestre, Polidoro, Sumbaly, Taylor, Silva, Hou, Wang, Hosseini, Chennabasappa, Singh, Bell, Kim, Edunov, Nie, Narang, Raparthy, Shen, Wan, Bhosale, Zhang, Vandenhende, Batra, Whitman, Sootla, Collot, Gururangan, Borodinsky, Herman, Fowler, Sheasha, Georgiou, Scialom, Speckbacher, Mihaylov, Xiao, Karn, Goswami, Gupta, Ramanathan, Kerkez, Gonguet, Do, Vogeti, Albiero, Petrovic, Chu, Xiong, Fu, Meers, Martinet, Wang, Wang, Tan, Xia, Xie, Jia, Wang, Goldschlag, Gaur, Babaei, Wen, Song, Zhang, Li, Mao, Coudert, Yan, Chen, Papakipos, Singh, Srivastava, Jain, Kelsey, Shajnfeld, Gangidi, Victoria, Goldstand, Menon, Sharma, Boesenberg,
  Baevski, Feinstein, Kallet, Sangani, Teo, Yunus, Lupu, Alvarado, Caples, Gu, Ho, Poulton, Ryan, Ramchandani, Dong, Franco, Goyal, Saraf, Chowdhury, Gabriel, Bharambe, Eisenman, Yazdan, James, Maurer, Leonhardi, Huang, Loyd, Paola, Paranjape, Liu, Wu, Ni, Hancock, Wasti, Spence, Stojkovic, Gamido, Montalvo, Parker, Burton, Mejia, Liu, Wang, Kim, Zhou, Hu, Chu, Cai, Tindal, Feichtenhofer, Gao, Civin, Beaty, Kreymer, Li, Adkins, Xu, Testuggine, David, Parikh, Liskovich, Foss, Wang, Le, Holland, Dowling, Jamil, Montgomery, Presani, Hahn, Wood, Le, Brinkman, Arcaute, Dunbar, Smothers, Sun, Kreuk, Tian, Kokkinos, Ozgenel, Caggioni, Kanayet, Seide, Florez, Schwarz, Badeer, Swee, Halpern, Herman, Sizov, Guangyi, Zhang, Lakshminarayanan, Inan, Shojanazeri, Zou, Wang, Zha, Habeeb, Rudolph, Suk, Aspegren, Goldman, Zhan, Damlaj, Molybog, Tufanov, Leontiadis, Veliche, Gat, Weissman, Geboski, Kohli, Lam, Asher, Gaya, Marcus, Tang, Chan, Zhen, Reizenstein, Teboul, Zhong, Jin, Yang, Cummings, Carvill, Shepard, McPhie,
  Torres, Ginsburg, Wang, Wu, U, Saxena, Khandelwal, Zand, Matosich, Veeraraghavan, Michelena, Li, Jagadeesh, Huang, Chawla, Huang, Chen, Garg, A, Silva, Bell, Zhang, Guo, Yu, Moshkovich, Wehrstedt, Khabsa, Avalani, Bhatt, Mankus, Hasson, Lennie, Reso, Groshev, Naumov, Lathi, Keneally, Liu, Seltzer, Valko, Restrepo, Patel, Vyatskov, Samvelyan, Clark, Macey, Wang, Hermoso, Metanat, Rastegari, Bansal, Santhanam, Parks, White, Bawa, Singhal, Egebo, Usunier, Mehta, Laptev, Dong, Cheng, Chernoguz, Hart, Salpekar, Kalinli, Kent, Parekh, Saab, Balaji, Rittner, Bontrager, Roux, Dollar, Zvyagina, Ratanchandani, Yuvraj, Liang, Alao, Rodriguez, Ayub, Murthy, Nayani, Mitra, Parthasarathy, Li, Hogan, Battey, Wang, Howes, Rinott, Mehta, Siby, Bondu, Datta, Chugh, Hunt, Dhillon, Sidorov, Pan, Mahajan, Verma, Yamamoto, Ramaswamy, Lindsay, Lindsay, Feng, Lin, Zha, Patil, Shankar, Zhang, Zhang, Wang, Agarwal, Sajuyigbe, Chintala, Max, Chen, Kehoe, Satterfield, Govindaprasad, Gupta, Deng, Cho, Virk, Subramanian, Choudhury,
  Goldman, Remez, Glaser, Best, Koehler, Robinson, Li, Zhang, Matthews, Chou, Shaked, Vontimitta, Ajayi, Montanez, Mohan, Kumar, Mangla, Ionescu, Poenaru, Mihailescu, Ivanov, Li, Wang, Jiang, Bouaziz, Constable, Tang, Wu, Wang, Wu, Gao, Kleinman, Chen, Hu, Jia, Qi, Li, Zhang, Zhang, Adi, Nam, Yu, Wang, Zhao, Hao, Qian, Li, He, Rait, DeVito, Rosnbrick, Wen, Yang, Zhao, and Ma}]{grattafiori2024llama3herdmodels}
Aaron Grattafiori, Abhimanyu Dubey, Abhinav Jauhri, Abhinav Pandey, Abhishek Kadian, Ahmad Al-Dahle, Aiesha Letman, Akhil Mathur, Alan Schelten, Alex Vaughan, Amy Yang, Angela Fan, Anirudh Goyal, Anthony Hartshorn, Aobo Yang, Archi Mitra, Archie Sravankumar, Artem Korenev, Arthur Hinsvark, and 542 others. 2024.
\newblock \href {https://arxiv.org/abs/2407.21783} {The llama 3 herd of models}.
\newblock \emph{Preprint}, arXiv:2407.21783.

\bibitem[{Han et~al.(2024)Han, Rao, Ettinger, Jiang, Lin, Lambert, Choi, and Dziri}]{han2024wildguardopenonestopmoderation}
Seungju Han, Kavel Rao, Allyson Ettinger, Liwei Jiang, Bill~Yuchen Lin, Nathan Lambert, Yejin Choi, and Nouha Dziri. 2024.
\newblock \href {https://arxiv.org/abs/2406.18495} {Wildguard: Open one-stop moderation tools for safety risks, jailbreaks, and refusals of llms}.
\newblock \emph{Preprint}, arXiv:2406.18495.

\bibitem[{Hoffmann et~al.(2022)Hoffmann, Borgeaud, Mensch, Buchatskaya, Cai, Rutherford, Casas, Hendricks, Welbl, Clark et~al.}]{hoffmann2022training}
Jordan Hoffmann, Sebastian Borgeaud, Arthur Mensch, Elena Buchatskaya, Trevor Cai, Eliza Rutherford, Diego de~Las Casas, Lisa~Anne Hendricks, Johannes Welbl, Aidan Clark, and 1 others. 2022.
\newblock Training compute-optimal large language models.
\newblock \emph{arXiv preprint arXiv:2203.15556}.

\bibitem[{Hu et~al.(2024)Hu, Piet, Zhao, Jiao, and Wagner}]{hu2024toxicitydetectionfree}
Zhanhao Hu, Julien Piet, Geng Zhao, Jiantao Jiao, and David Wagner. 2024.
\newblock \href {https://arxiv.org/abs/2405.18822} {Toxicity detection for free}.
\newblock \emph{Preprint}, arXiv:2405.18822.

\bibitem[{Huang et~al.(2025)Huang, Hu, Ilhan, Tekin, Yahn, Xu, and Liu}]{huang2025safetytaxsafetyalignment}
Tiansheng Huang, Sihao Hu, Fatih Ilhan, Selim~Furkan Tekin, Zachary Yahn, Yichang Xu, and Ling Liu. 2025.
\newblock \href {https://arxiv.org/abs/2503.00555} {Safety tax: Safety alignment makes your large reasoning models less reasonable}.
\newblock \emph{Preprint}, arXiv:2503.00555.

\bibitem[{Inan et~al.(2023)Inan, Upasani, Chi, Rungta, Iyer, Mao, Tontchev, Hu, Fuller, Testuggine, and Khabsa}]{inan2023llamaguardllmbasedinputoutput}
Hakan Inan, Kartikeya Upasani, Jianfeng Chi, Rashi Rungta, Krithika Iyer, Yuning Mao, Michael Tontchev, Qing Hu, Brian Fuller, Davide Testuggine, and Madian Khabsa. 2023.
\newblock \href {https://arxiv.org/abs/2312.06674} {Llama guard: Llm-based input-output safeguard for human-ai conversations}.
\newblock \emph{Preprint}, arXiv:2312.06674.

\bibitem[{Jawahar et~al.(2019)Jawahar, Sagot, and Seddah}]{Jawahar2019}
Ganesh Jawahar, Beno{\^{i}}t Sagot, and Djam{\'{e}} Seddah. 2019.
\newblock \href {https://doi.org/10.18653/v1/P19-1356} {{What Does BERT Learn about the Structure of Language?}}
\newblock In \emph{Proceedings of the 57th Annual Meeting of the Association for Computational Linguistics}, pages 3651--3657, Stroudsburg, PA, USA. Association for Computational Linguistics.

\bibitem[{Kaplan et~al.(2020)Kaplan, McCandlish, Henighan, Brown, Chess, Child, Gray, Radford, Wu, and Amodei}]{kaplan2020scaling}
Jared Kaplan, Sam McCandlish, Tom Henighan, Tom~B. Brown, Benjamin Chess, Rewon Child, Scott Gray, Alec Radford, Jeffrey Wu, and Dario Amodei. 2020.
\newblock \href {https://arxiv.org/abs/2001.08361} {Scaling laws for neural language models}.
\newblock \emph{arXiv preprint arXiv:2001.08361}.

\bibitem[{Li et~al.(2025)Li, Wei, Zhang, and Zhang}]{li2025eagle3scalinginferenceacceleration}
Yuhui Li, Fangyun Wei, Chao Zhang, and Hongyang Zhang. 2025.
\newblock \href {https://arxiv.org/abs/2503.01840} {Eagle-3: Scaling up inference acceleration of large language models via training-time test}.
\newblock \emph{Preprint}, arXiv:2503.01840.

\bibitem[{Lin et~al.(2023)Lin, Wang, Tong, Wang, Guo, Wang, and Shang}]{lin-etal-2023-toxicchat}
Zi~Lin, Zihan Wang, Yongqi Tong, Yangkun Wang, Yuxin Guo, Yujia Wang, and Jingbo Shang. 2023.
\newblock \href {https://doi.org/10.18653/v1/2023.findings-emnlp.311} {{T}oxic{C}hat: Unveiling hidden challenges of toxicity detection in real-world user-{AI} conversation}.
\newblock In \emph{Findings of the Association for Computational Linguistics: EMNLP 2023}, pages 4694--4702, Singapore. Association for Computational Linguistics.

\bibitem[{Maas et~al.(2011)Maas, Daly, Pham, Huang, Ng, and Potts}]{maas2011learning}
Andrew Maas, Raymond~E Daly, Peter~T Pham, Dan Huang, Andrew~Y Ng, and Christopher Potts. 2011.
\newblock Learning word vectors for sentiment analysis.
\newblock In \emph{Proceedings of the 49th annual meeting of the association for computational linguistics: Human language technologies}, pages 142--150.

\bibitem[{OpenAI et~al.(2025)OpenAI, :, Agarwal, Ahmad, Ai, Altman, Applebaum, Arbus, Arora, Bai, Baker, Bao, Barak, Bennett, Bertao, Brett, Brevdo, Brockman, Bubeck, Chang, Chen, Chen, Cheung, Clark, Cook, Dukhan, Dvorak, Fives, Fomenko, Garipov, Georgiev, Glaese, Gogineni, Goucher, Gross, Guzman, Hallman, Hehir, Heidecke, Helyar, Hu, Huet, Huh, Jain, Johnson, Koch, Kofman, Kundel, Kwon, Kyrylov, Le, Leclerc, Lennon, Lessans, Lezcano-Casado, Li, Li, Lin, Liss, Lily, Liu, Liu, Lu, Lu, Martinovic, McCallum, McGrath, McKinney, McLaughlin, Mei, Mostovoy, Mu, Myles, Neitz, Nichol, Pachocki, Paino, Palmie, Pantuliano, Parascandolo, Park, Pathak, Paz, Peran, Pimenov, Pokrass, Proehl, Qiu, Raila, Raso, Ren, Richardson, Robinson, Rotsted, Salman, Sanjeev, Schwarzer, Sculley, Sikchi, Simon, Singhal, Song, Stuckey, Sun, Tillet, Toizer, Tsimpourlas, Vyas, Wallace, Wang, Wang, Watkins, Weil, Wendling, Whinnery, Whitney, Wong, Yang, Yang, Yasunaga, Ying, Zaremba, Zhan, Zhang, Zhang, Zhang, and
  Zhao}]{openai2025gptoss120bgptoss20bmodel}
OpenAI, :, Sandhini Agarwal, Lama Ahmad, Jason Ai, Sam Altman, Andy Applebaum, Edwin Arbus, Rahul~K. Arora, Yu~Bai, Bowen Baker, Haiming Bao, Boaz Barak, Ally Bennett, Tyler Bertao, Nivedita Brett, Eugene Brevdo, Greg Brockman, Sebastien Bubeck, and 108 others. 2025.
\newblock \href {https://arxiv.org/abs/2508.10925} {gpt-oss-120b \& gpt-oss-20b model card}.
\newblock \emph{Preprint}, arXiv:2508.10925.

\bibitem[{Palma et~al.(2025)Palma, Bellis, Servedio, Anelli, Narducci, and Noia}]{dipalma2025llamasfeelingstoounveiling}
Dario~Di Palma, Alessandro~De Bellis, Giovanni Servedio, Vito~Walter Anelli, Fedelucio Narducci, and Tommaso~Di Noia. 2025.
\newblock \href {https://arxiv.org/abs/2505.16491} {Llamas have feelings too: Unveiling sentiment and emotion representations in llama models through probing}.
\newblock \emph{Preprint}, arXiv:2505.16491.

\bibitem[{Rogers et~al.(2020)Rogers, Kovaleva, and Rumshisky}]{rogers2020primer}
Anna Rogers, Olga Kovaleva, and Anna Rumshisky. 2020.
\newblock \href {https://doi.org/10.1162/tacl_a_00349} {A primer in {BERT}ology: What we know about how {BERT} works}.
\newblock \emph{Transactions of the Association for Computational Linguistics}, 8:842--866.

\bibitem[{Saravia et~al.(2018)Saravia, Liu, Huang, Wu, and Chen}]{saravia-etal-2018-carer}
Elvis Saravia, Hsien-Chi~Toby Liu, Yen-Hao Huang, Junlin Wu, and Yi-Shin Chen. 2018.
\newblock \href {https://doi.org/10.18653/v1/D18-1404} {{CARER}: Contextualized affect representations for emotion recognition}.
\newblock In \emph{Proceedings of the 2018 Conference on Empirical Methods in Natural Language Processing}, pages 3687--3697, Brussels, Belgium. Association for Computational Linguistics.

\bibitem[{Socher et~al.(2013)Socher, Perelygin, Wu, Chuang, Manning, Ng, and Potts}]{socher2013recursive}
Richard Socher, Alex Perelygin, Jean Wu, Jason Chuang, Christopher~D Manning, Andrew~Y Ng, and Christopher Potts. 2013.
\newblock Recursive deep models for semantic compositionality over a sentiment treebank.
\newblock In \emph{Proceedings of the 2013 conference on empirical methods in natural language processing}, pages 1631--1642.

\bibitem[{Tenney et~al.(2019)Tenney, Das, and Pavlick}]{tenney2019bertrediscoversclassicalnlp}
Ian Tenney, Dipanjan Das, and Ellie Pavlick. 2019.
\newblock \href {https://arxiv.org/abs/1905.05950} {Bert rediscovers the classical nlp pipeline}.
\newblock \emph{Preprint}, arXiv:1905.05950.

\bibitem[{Vaswani et~al.(2017)Vaswani, Shazeer, Parmar, Uszkoreit, Jones, Gomez, Kaiser, and Polosukhin}]{NIPS2017_3f5ee243}
Ashish Vaswani, Noam Shazeer, Niki Parmar, Jakob Uszkoreit, Llion Jones, Aidan~N Gomez, \L~ukasz Kaiser, and Illia Polosukhin. 2017.
\newblock \href {https://proceedings.neurips.cc/paper_files/paper/2017/file/3f5ee243547dee91fbd053c1c4a845aa-Paper.pdf} {Attention is all you need}.
\newblock In \emph{Advances in Neural Information Processing Systems}, volume~30. Curran Associates, Inc.

\bibitem[{Verma et~al.(2025)Verma, Hines, Bilmes, Siska, Zettlemoyer, Gonen, and Singh}]{verma2025omniguardefficientapproachai}
Sahil Verma, Keegan Hines, Jeff Bilmes, Charlotte Siska, Luke Zettlemoyer, Hila Gonen, and Chandan Singh. 2025.
\newblock \href {https://arxiv.org/abs/2505.23856} {Omniguard: An efficient approach for ai safety moderation across modalities}.
\newblock \emph{Preprint}, arXiv:2505.23856.

\bibitem[{Xuan et~al.(2025)Xuan, Mao, Chen, Zhang, Dong, and Zhou}]{xuan-etal-2025-shieldhead}
Zitao Xuan, Xiaofeng Mao, Da~Chen, Xin Zhang, Yuhan Dong, and Jun Zhou. 2025.
\newblock \href {https://doi.org/10.18653/v1/2025.findings-acl.932} {{S}hield{H}ead: Decoding-time safeguard for large language models}.
\newblock In \emph{Findings of the Association for Computational Linguistics: ACL 2025}, pages 18129--18143, Vienna, Austria. Association for Computational Linguistics.

\bibitem[{Yang et~al.(2025)Yang, Li, Yang, Zhang, Hui, Zheng, Yu, Gao, Huang, Lv, Zheng, Liu, Zhou, Huang, Hu, Ge, Wei, Lin, Tang, Yang, Tu, Zhang, Yang, Yang, Zhou, Zhou, Lin, Dang, Bao, Yang, Yu, Deng, Li, Xue, Li, Zhang, Wang, Zhu, Men, Gao, Liu, Luo, Li, Tang, Yin, Ren, Wang, Zhang, Ren, Fan, Su, Zhang, Zhang, Wan, Liu, Wang, Cui, Zhang, Zhou, and Qiu}]{yang2025qwen3technicalreport}
An~Yang, Anfeng Li, Baosong Yang, Beichen Zhang, Binyuan Hui, Bo~Zheng, Bowen Yu, Chang Gao, Chengen Huang, Chenxu Lv, Chujie Zheng, Dayiheng Liu, Fan Zhou, Fei Huang, Feng Hu, Hao Ge, Haoran Wei, Huan Lin, Jialong Tang, and 41 others. 2025.
\newblock \href {https://arxiv.org/abs/2505.09388} {Qwen3 technical report}.
\newblock \emph{Preprint}, arXiv:2505.09388.

\bibitem[{Zheng et~al.(2023)Zheng, Chiang, Sheng, Zhuang, Wu, Zhuang, Lin, Li, Li, Xing et~al.}]{zheng2023judging}
Lianmin Zheng, Wei-Lin Chiang, Ying Sheng, Siyuan Zhuang, Zhanghao Wu, Yonghao Zhuang, Zi~Lin, Zhuohan Li, Dacheng Li, Eric Xing, and 1 others. 2023.
\newblock Judging llm-as-a-judge with mt-bench and chatbot arena.
\newblock \emph{Advances in neural information processing systems}, 36:46595--46623.

\bibitem[{Zhou et~al.(2024)Zhou, Yu, Zhang, Xu, Huang, and Li}]{zhou-etal-2024-alignment}
Zhenhong Zhou, Haiyang Yu, Xinghua Zhang, Rongwu Xu, Fei Huang, and Yongbin Li. 2024.
\newblock \href {https://doi.org/10.18653/v1/2024.findings-emnlp.139} {How alignment and jailbreak work: Explain {LLM} safety through intermediate hidden states}.
\newblock In \emph{Findings of the Association for Computational Linguistics: EMNLP 2024}, pages 2461--2488, Miami, Florida, USA. Association for Computational Linguistics.

\end{thebibliography}
\newpage
\appendix

\section{Hyperparameter Configuration and Sensitivity}
\label{app:hyperparameter-search}
\begin{figure*}[h]
    \centering
    \includegraphics[width=1\linewidth]{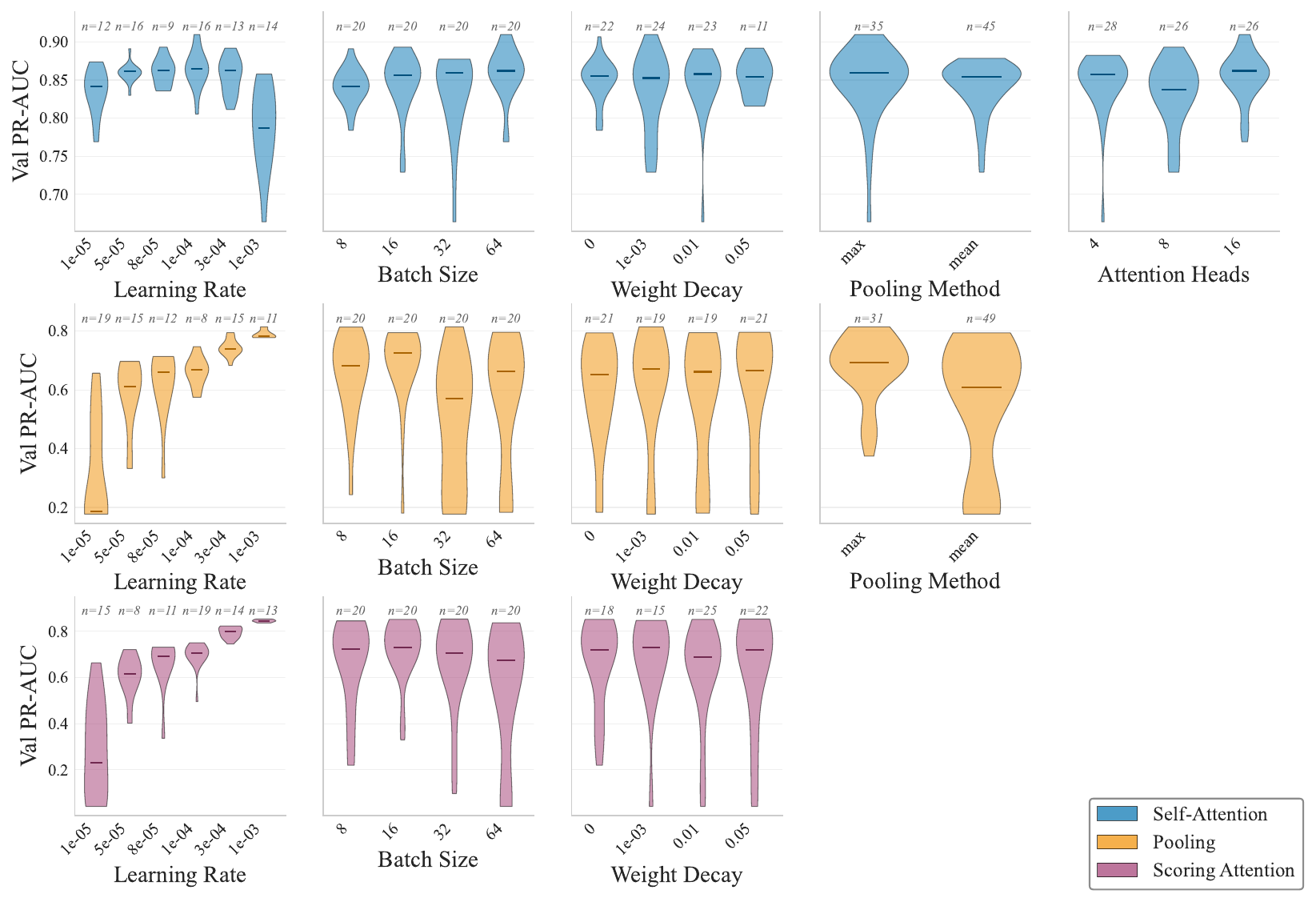}
    \caption{Sensitivity of PR-AUC to hyperparameter choices on ToxicChat across ${\sim}100$ configurations. Each row corresponds to an aggregation method; horizontal lines denote medians. Self-Attention sustains high PR-AUC (${\sim}0.75$–$0.90$) with low variance across settings. Pooling and Scoring Attention are highly sensitive to learning rate, spanning ${\sim}0.2$–$0.9$.}

    \label{fig:hyperparam_effects}
\end{figure*}

We conducted hyperparameter search for each dataset and aggregation mechanism. A key advantage of our lightweight probes is that this scale of experimentation (approximately 100 configurations per dataset) would be prohibitive when fine-tuning multi-billion parameter guard models. Beyond finding optimal configurations, this enables systematic sensitivity analysis: characterizing how performance varies across the design space.

All search was performed using a train/validation split, with the test set held out for final evaluation. We employed early stopping based on validation F1 score for safety datasets and validation accuracy for sentiment datasets.

\subsection{Search Ranges}

Table~\ref{tab:hyperparam_ranges} summarizes the hyperparameter ranges explored across all experiments.

\begin{table}[!t]
\centering
\small
\begin{tabular}{ll}
\toprule
\textbf{Hyperparameter} & \textbf{Search Range} \\
\midrule
Learning rate & $[10^{-5}, 10^{-3}]$ \\
Batch size & $\{8, 16, 32, 64\}$ \\
Max epochs & 10 (with early stopping) \\
Optimizer & AdamW ($\beta_1{=}0.9$, $\beta_2{=}0.999$) \\
Weight decay & $[0, 0.05]$ \\
LR scheduler & Cosine annealing \\
Pooling method & $\{\text{mean}, \text{max}\}$ \\
Attention heads & $\{4, 8, 16\}^{\dagger}$ \\
Downcast factor & $\{4, 8, 16, 32, 64\}^{\ddagger}$ \\
\bottomrule
\end{tabular}
\caption{Hyperparameter search ranges for the primary backbone
(Llama-3.2-3B, $d{=}3072$). $\dagger$Adjusted per backbone for
divisibility: $\{1,3,9,15\}$ for GPT-OSS-20B ($d{=}2880$),
$\{1,2,4,8\}$ for Qwen3-30B-A3B ($d{=}2048$). $\ddagger$Fixed to 32
for GPT-OSS-20B and Qwen3-30B-A3B.}
\label{tab:hyperparam_ranges}
\end{table}

\subsection{Experimental Setup}
\label{sec:experimental-setup}

All probe training was conducted on a single NVIDIA RTX 3090 GPU (24GB VRAM) with 96GB system RAM. For the additional backbones (GPT-OSS-20B and Qwen3-30B-A3B), hidden states were pre-extracted on an NVIDIA A100 80GB GPU and transferred for local probe training. To reduce peak GPU memory during training, we pre-extracted and cached hidden states from the frozen LLM before training probe classifiers, allowing larger batch sizes under VRAM constraints.

\subsection{Sensitivity Analysis}

Figure~\ref{fig:hyperparam_effects} shows PR-AUC sensitivity to hyperparameter choices on ToxicChat, revealing that learning rate is the dominant factor for Pooling and Scoring Attention, while Self-Attention is comparatively robust across settings.

\begin{table*}[!t]
\centering
\small
\begin{tabular}{lcccccc}
\toprule
\textbf{Dataset} & \textbf{Downcast} & \textbf{Accuracy (\%)} & \textbf{F1 (\%)} & \textbf{ROC AUC} & \textbf{PR AUC} & \textbf{Params (M)} \\
\midrule
\multirow{5}{*}{ToxicChat} 
& 4  & 96.06 ± 0.33 & 84.13 ± 0.86 & \textbf{0.983} ± 0.002 & \textbf{0.906} ± 0.008 & 283 \\
& 8  & 95.89 ± 0.09 & 82.93 ± 1.40 & 0.981 ± 0.002 & 0.902 ± 0.012 & 142 \\
& 16 & 95.57 ± 0.20 & 83.02 ± 0.39 & 0.981 ± 0.001 & 0.889 ± 0.004 & 71 \\
& 32 & \textbf{96.13} ± 0.19 & \textbf{84.51} ± 0.43 & 0.981 ± 0.001 & 0.898 ± 0.006 & 35 \\
& 64 & 95.91 ± 0.12 & 82.80 ± 1.36 & 0.981 ± 0.001 & 0.904 ± 0.002 & 18 \\
\midrule
\multirow{5}{*}{SST-2} 
& 4  & 89.88 ± 5.74 & 90.95 ± 4.57 & 0.990 ± 0.001 & 0.988 ± 0.003 & 283 \\
& 8  & \textbf{95.44} ± 0.22 & \textbf{95.52} ± 0.20 & \textbf{0.992} ± 0.001 & \textbf{0.992} ± 0.001 & 142 \\
& 16 & 91.95 ± 1.47 & 92.51 ± 1.24 & \textbf{0.992} ± 0.000 & \textbf{0.992} ± 0.000 & 71 \\
& 32 & 95.39 ± 0.32 & 95.31 ± 0.33 & \textbf{0.992} ± 0.001 & \textbf{0.992} ± 0.001 & 35 \\
& 64 & 93.70 ± 2.39 & 94.01 ± 2.04 & 0.991 ± 0.000 & 0.991 ± 0.000 & 18 \\
\bottomrule
\end{tabular}
\caption{Effect of attention downcasting on ToxicChat and SST-2 performance. Results show mean ± std across 3 runs. Lower downcasting factors (larger $d_{\text{inner}}$) increase parameter count. Downcast 32 is optimal for ToxicChat, while downcast 8 achieves best results on SST-2.}
\label{tab:downcasting-ablation}
\end{table*}

\FloatBarrier
\section{Ablation Study on Attention Downcasting}
We evaluate the effect of dimension downcasting on the multi-head self-attention aggregation mechanism using the ToxicChat and SST2 datasets. The results are shown in Table \ref{tab:downcasting-ablation}. The downcasting factor determines the inner dimension $d_{\text{inner}} = d/\text{factor}$ for the QKV projections, where $d$ is the hidden dimension ($d = 3{,}072$ for Llama-3.2-3B).

All experiments use the same hyperparameters configuration from our main results, with only the downcast parameter being altered.

\begin{figure*}[!t]
    \centering
    \includegraphics[width=1\linewidth]{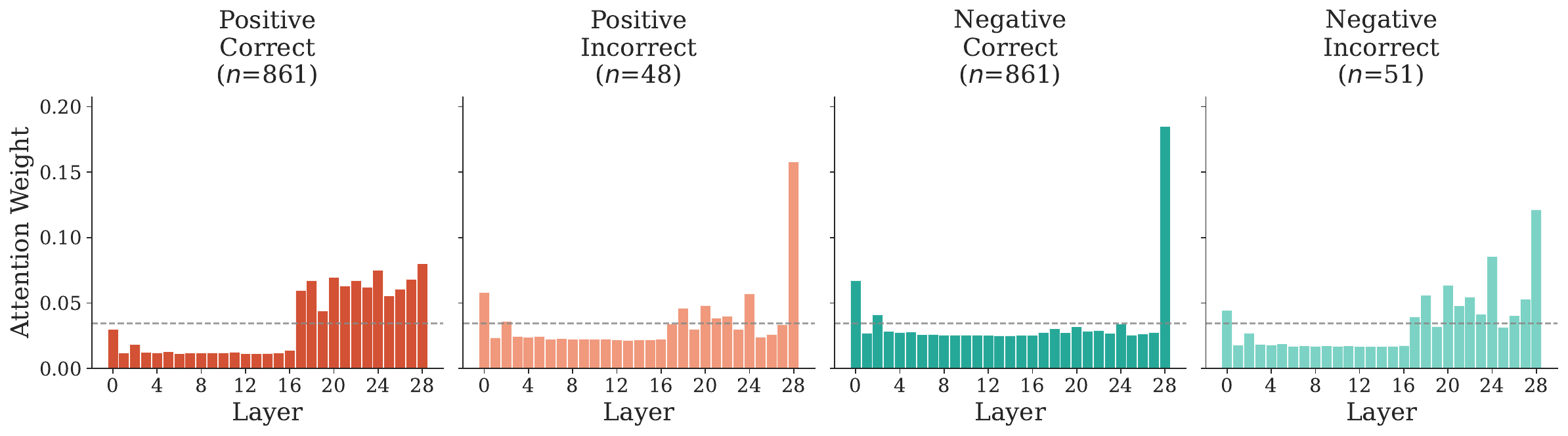}
    \caption{Layer-wise post-softmax attention weights on SST-2, stratified by ground-truth label (Positive vs. Negative) and prediction correctness. Similar to toxicity detection (Figure \ref{fig:layer-attention-tc}), correctly classified samples exhibit distinct class-conditional attention patterns: positive sentiment concentrates on later intermediate layers (L17--L28), while negative sentiment shows concentrated attention weights on the embedding and final layers (L0, L28).}
    \label{fig:layer-attention-sst2}
\end{figure*}

\FloatBarrier
\section{Layer Attention on SST-2}
\label{app:attention-analysis}
Figure \ref{fig:layer-attention-sst2} shows layer-wise attention patterns for sentiment classification on SST-2, paralleling our toxicity analysis. The task-specific attention distributions further validate that different features related to sentiment classification emerge at different layer depths, supporting our learned aggregation approach over single layer selection.

\end{document}